\documentclass[10pt,twocolumn,letterpaper]{article}

\usepackage{cvpr}
\usepackage{times}
\usepackage{epsfig}
\usepackage{graphicx}
\usepackage{amsmath}
\usepackage{amssymb}

\cvprfinalcopy 

\def\cvprPaperID{****} 

\begin{document}

\title{Learning Local Features with Context Aggregation for Visual Localization}
\author{Siyu Hong\begin{math}^{1,3}\end{math} \quad Kunhong Li\begin{math}^1\end{math} \quad Yongcong Zhang\begin{math}^1\end{math} \quad Zhiheng Fu\begin{math}^2\end{math} \quad Mengyi Liu\begin{math}^3\end{math} \quad Yulan Guo\begin{math}^1\end{math} \\
	\begin{math}{}^1\end{math}Sun Yat-sen University \quad \begin{math}{}^2\end{math}The University of Western Australia \quad \begin{math}{}^3\end{math}Alibaba Group, Beijing, China\\
	{\tt\small \{hongsy3,likh25,zhangyc6\}@mail2.sysu.edu.cn \quad zhihengfu92128@gmail.com}\\
	{\tt\small suqing.lmy@alibaba-inc.com \quad guoyulan@sysu.edu.cn }
}
\maketitle
\begin{abstract}
    Keypoint detection and description is fundamental yet important in many vision applications. Most existing methods use detect-then-describe or detect-and-describe strategy to learn local features without considering their context information. Consequently, it is challenging for these methods to learn robust local features. In this paper, we focus on the fusion of low-level textual information and high-level semantic context information to improve the discrimitiveness of local features. Specifically, we first estimate a score map to represent the distribution of potential keypoints according to the quality of descriptors of all pixels. Then, we extract and aggregate multi-scale high-level semantic features based by the guidance of the score map. Finally, the low-level local features and high-level semantic features are fused and refined using a residual module. Experiments on the challenging local feature benchmark dataset demonstrate that our method achieves the state-of-the-art performance in the local feature challenge of the visual localization benchmark.
\end{abstract}

\section{Introduction}
Visual localization is an important research topic in computer vision since it plays a critical role in many tasks such as loop closure detection and re-localization. Three approaches are commonly used to solve this problem, including 3D structure-based methods~\cite{sattler2016efficient}\cite{liu2017efficient}\cite{zeisl2015camera}\cite{svarm2016city}, 2D image-based methods~\cite{arandjelovic2016netvlad}\cite{lowry2015visual}\cite{sattler2016large}\cite{torii201524} and learning-based methods~\cite{chen2017deep}\cite{cao2013graph}\cite{kendall2015posenet}. Among these methods, 3D structure-based methods are more accurate than other methods.

For a 3D structure-based method, correspondences are first generated between 2D local features in the query image and 3D points constructed by dense Structure-from-Motion(SfM) models. The 6DoF camera pose is then recovered by PnP. Consequently, local features learning plays a vital role in visual localization. 

Traditional local feature learning methods commonly detect keypoints in the first stage and then describe each keypoint as a local descriptor. In general, a good local feature should have three types of properties, i) High repeatability: robust local features can be detected in different images. ii) High reliability: each local feature in an image should be descriminative. iii) High efficiency: the number of local features is much less than the number of pixels. Aiming for these properties, many handcrafted local features have been presented. For keypoint detection, corners~\cite{harris1988combined} or blobs~\cite{lowe2004distinctive} are considered as low-level image information. Meanwhile, numerous handcrafted descriptors (such as histograms of gradient (HOG)~\cite{dalal2005histograms} and its variants~\cite{bay2006surf}) have also been proposed. 

However, handcrafted methods are limited by the prior knowledge and easily influenced by the change of illumination, weather and season. With the development of deep learning, recent methods jointly learn keypoint detectors and descriptors in an end-to-end manner. Some methods learn the correspondences from dense SfM models produced by Colmap~\cite{schonberger2016structure}. SuperPoint~\cite{detone2018superpoint} learns the simple homography transformation using an artifact dataset. Sarlin et al.~\cite{sarlin2019coarse} proposed highly robust correspondances using teacher-student networks in HFNet. Revaud et al.~\cite{revaud2019r2d2} proposed EpicFlow~\cite{revaud2015epicflow} to produce semi-sparse matches and then learn correspondences. However, high-level semantic information of the image patches around local features is not fully considered in these. 

Besides, visual localization still suffers from day-night and season variations. It is difficult to recover camera poses due to the weakness of local features when images are acquired across a long time. In contrast,  high-level semantic information of an image is more robust and invariant than local features under these cases. To achieve robust performance, it is therefore reasonable to fuse low-level textural information with high-level semantic information.

In this paper, we propose a robust method to learn and combine high-level semantic information with its corresponding low-level textual features. Since high-level semantic information is constant under different conditions, the repeatability and reliability of the fused local features can be finally improved. Following the previous work R2D2~\cite{revaud2019r2d2}, we use the repeatability map and the reliability map to generate the score map. Then, we use the score map to mask the semantic features produced by HRNet~\cite{sun2019deep}. This local-guided module is employed to choose semantic features adaptively. Next, we apply a multi-scale aggregation module to aggregate semantic features in a specific neighborhood. Finally, we design a residual module to refine the fused features.

Compared to recent local feature learning approaches~\cite{luo2019contextdesc}\cite{dusmanu2019d2}\cite{revaud2019r2d2}\cite{luo2020aslfeat}, our work makes the following contributions: 1) Different from Contextdesc~\cite{luo2019contextdesc}, we fuse semantic information with local features before keypoint detection to obtain robust keypoints and descriptors. 2) We design a residual module to refine low-level textual features automatically. 3) Compared to R2D2, no additional constraint is required by our work.
\section{Related Work}
{\bf Localization Benchmarks.} KITTI~\cite{carlevaris2016university}, TorontoCity~\cite{wang2016torontocity} and the Michigan North Campus Long-Term (NCLT)~\cite{geiger2013vision} are commonly used visual localization benchmarks. However, these benchmarks do not exhibit all complex scenes in the real world. To hanlde this problem, Sattler et al.~\cite{sattler2018benchmarking} introduced the Aachen Day-Night dataset for local feature and device challenges. The Aachen Day-Night dataset provides several types of challenging scenes, including day-night changes and illumination changes (dawn/noon). Besides, they pproposed the RobotCar Seasons Dataset and the CMU Seasons Dataset for autonomous vehicles challenges, including weather (dust/sun/rain/snow) and seasonal (winter/summer) variations. In this paper, we focus on local feature learning. Therefore, we evaluate our method on the Aachen Day-Night dataset.

{\bf Joint Local Feature Learning.} Joint learning of feature detectors and descriptors has attracted increasing attention due to its computational efficiency. Yi et al.~\cite{yi2016lift} used three subnetworks to predict keypoints, their corresponding orientations and descriptors. Besides, they replaced the Non-Maximum Suppression (NMS) with the soft argmax function. 
DeTone et al.~\cite{detone2018superpoint} used artificial homography transformations to generating image pairs as the training data, which were used to learn point-wise pixel level correspondences. Yuki et al.~\cite{ono2018lf} generated the relative pose and corresponding depth maps of image pairs by SfM. Then, they proposed LFNet to minimize the difference between image pairs. Recently, R2D2~\cite{revaud2019r2d2} simultaneously estimated descriptors, reliability map and repeatability map. Reliability map and repeatability map are further used to detect keypoints. In D2-Net~\cite{dusmanu2019d2}, a describe-and-detect approach is proposed by generating descriptors and keypoints from the same feature maps. However, this method is hard to detect accurate keypoints learned from low-resolution feature maps. Based on D2-Net, Deformable Convolutional Network (DCN)~\cite{dai2017deformable}\cite{zhu2019deformable} is used in ASLFeat~\cite{luo2020aslfeat} to model geometric variations. Besides, multi-level keypoint detection is integrated into an end-to-end network in ASLFeat. Although dynamic receptive filed is learned 
by DCN, it still works on low-level textural information. In Contextdesc~\cite{luo2019contextdesc}, context information is leveraged to augment local descriptors after keypoint detection. In contrast, we aggregate high-level semantic features with local features before keypoint detection. Consequently, the proposed architecture achieves better performance on both keypoint detection and descriptor learning.

\section{Methods}

The backbone architecture in this work consists of two parts: 1) a high-resolution network (HRNet)~\cite{sun2019deep} that extracts high-level semantic features, and 2) a R2D2 network~\cite{revaud2019r2d2} that jointly learns local features, reliability maps and repeatability maps.

\subsection{Prerequisites}
{\bf High-resolution network (HRNet)~\cite{sun2019deep}}  is able to maintain a high resolution without recovering the resolution through a low-to-high process. Different from other existing methods, HRNet can boost the high-resolution representation by fusing low-resolution representations at different scales. Consequently, better high-level semantic features can be produced for local feature learning. 

{\bf R2D2~\cite{revaud2019r2d2} } enables repeatability and reliability prediction for descriptors. Based on L2Net~\cite{tian2017l2}, the last $8 \times 8$ convolutional layer is replaced by three successive $2 \times 2$ convolutional layers to improve computational efficiency. Based on the output of L2Net, an L2-normalization layer is used in R2D2 to obtain dense feature descriptors. Meanwhile, a reliability map and a repeatability map are predicted in R2D2 using an element-wise square operation, a $1 \times 1$ convolutional layer and a softmax operation. In addition, a self-supervised loss is introduced in R2D2 to learn repeatable and reliable keypoints and descriptors.

\begin{figure*}
	\begin{center}
		\includegraphics[width=0.9\textwidth]{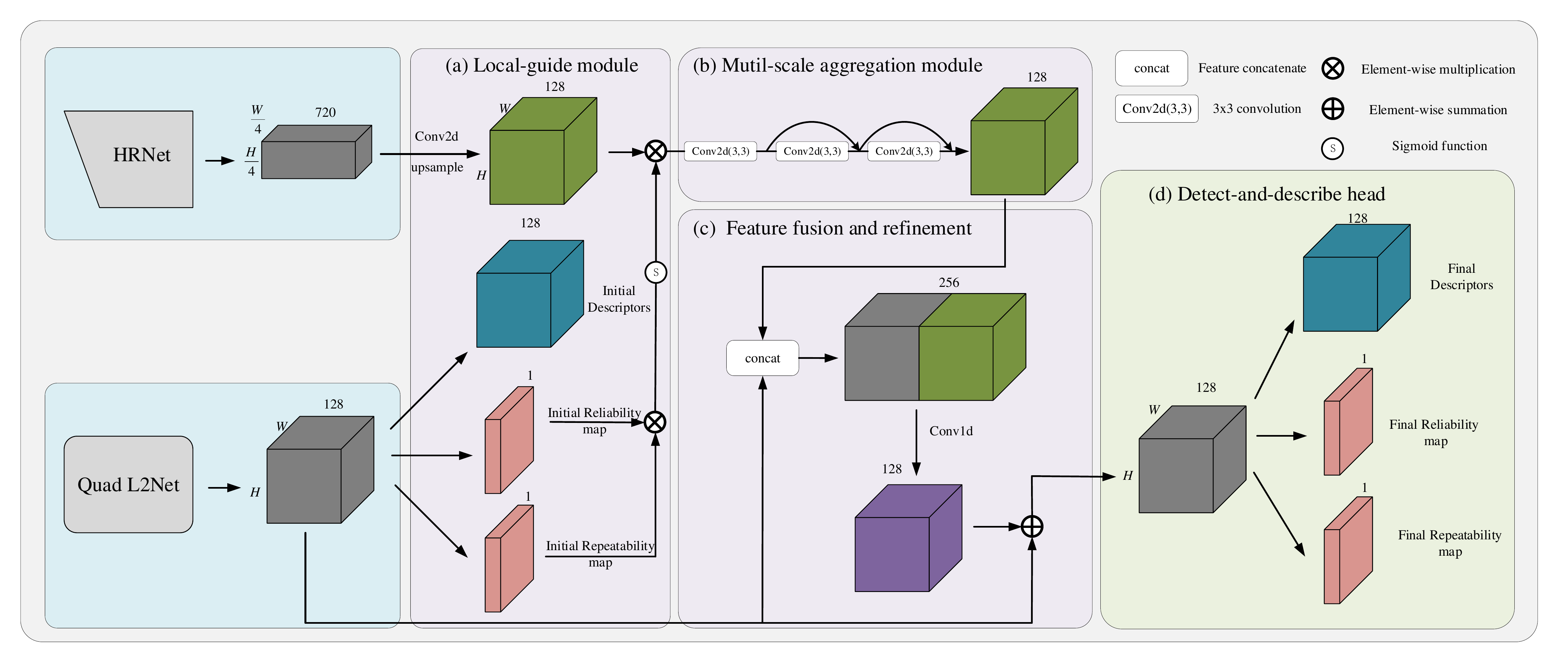}
	\end{center}
	\caption{An overview of the proposed method, which consists of a local-guided module, a multi-scale aggregation module, and feature fusion and refinement.}
	\label{fig:net}
\end{figure*}
\subsection{Network Architecture}
The network architecture is illustrated in Figure~\ref{fig:net}. Following R2D2~\cite{revaud2019r2d2}, we replace the last $8 \times 8$ convolution of L2-Net by three $3 \times 3$ convolutions, resulting in a $128$-dimension tensor. Then, we predict initial descriptors, initial repeatability and reliability maps using the tensor. Next, the initial repeatability and reliability maps are used to obtain the score map by element-wise multiply operation. To extract high-level semantic features, we apply the HRNet-W48 model. Then, we design a local-guided module using the score map to obtain the selective semantic features. Besides, we introduce a multi-scale aggregation module to aggregate the context information of the selective semantic features. The residual of low-level textual features can be obtained by a $1 \times 1$ convolution. After fusing the low-level textual features with the high-level semantic features, we use the same module as R2D2 to learn the final prediction. 

\subsection{Local-guided module}
To extract useful information from high-level semantic features, we use repeatability map and reliability map as the guidance, as shown in Figure~\ref{fig:net}. Following R2D2~\cite{revaud2019r2d2}, we perform element-wise multiplication between the repeatability map and the reliability map to obtain score maps during keypoint detection. Then, we use a sigmoid function to normalize the score map into a probability map. Each value in the probability map represents the probability of becoming a potential keypoint for a pixel.
\begin{equation}
{S}_{ij} = Sigmoid({Reliability}_{ij} \otimes {Repeatability}_{ij})
\end{equation}
where $i,j$ represent the position of pixel $(i,j)$. Then, we use this score map to choose the high-level semantic features by an element-wise multiplication operation. Finally, the selective semantic features is obtained. 

\subsection{Multi-scale aggregation module}
In Contextdesc~\cite{luo2019contextdesc}, descriptors are leveraged by the contextual high-level semantic features. We aggregate selective semantic features before their fusion with low-level textual features, as shown in Figure~\ref{fig:net}(b). Specifically, we use three successive $3 \times 3$ convolutions to aggregate the contextual information from $3 \times 3$, $5 \times 5$ and $7 \times 7$ fields. Short 
connections between every two consecutive layers are used for all convolutional layers in this module. Consequently, more information can be absorbed from a small field by the multi-scale aggregation module.

\subsection{Feature fusion and refinement}
The  feature map produced by the multi-scale aggregation module is considered as the residual of low-level textual features. Specifically, the output of the multi-scale aggregation module is first fused with the low-level textual features by concatenation. Then, channel aggregation is conducted on the concatenated feature map by a $1 \times 1$ convolution. Next, a refined feature map is obtained by an element-wise add operation between the low-level textual features and the output of the channel aggregation. Finally, the same head of R2D2 is used to predict the final descriptors, repeatability map and the reliability map.

\section{Experimental Evaluation}
We evaluate our method on the Aachen Day-Night dataset~\cite{sattler2018benchmarking}.

\subsection{Implementation details}

{\bf Localization Pipeline.} We follow the pipeline of Colmap~\cite{schonberger2016structure} to evaluate local features. First, our local features are imported into the database to generate a SfM model. Mutal NN matcher is used for feature matching. Then, query images are registered to this model. We follow Colmap to conduct the remaining steps, resulting in a 6DoF camera pose for each query image. The predicted 6DoF camera pose is then evaluated with three error tolerances $(0.5m, 2^{\circ})$ / $(1m, 5^{\circ})$ / $(5m, 10^{\circ})$, where the first number in each parentheses represents the position error while the second number represents the orientation error.

{\bf The Aachen Day-Night Dataset.} It contains 4,479 reference images and 369 query images acquired in Aachen, Germany, containing day-night variations. For the local feature challenge, we only predicted 6DoF camera poses of $98$ query images.

{\bf Training Details.} We used the Oxford and Paris retrieval dataset~\cite{radenovic2018revisiting} and some pairs from the Aachen Day-Night dataset~\cite{sattler2018benchmarking} as the training data. We trained the network for 25 epoches, using Adam with a batch size of 4, a learning rate of 0.001 and a weight decay of 0.0005. HRNet was pre-trained on the Cityscapes dataset~\cite{cordts2016cityscapes}. 

{\bf Loss Function.} Given two images $I$ and $I'$ of the same scene, their ground-truth correspondences  $U \in \mathbb{R}^{H \times W \times 2}$, and their repeatability maps $S$ and $S'$, a heatmap $S'_U$ from image $I'$ can be warped according to $U$. A set of overlapping patches $\mathcal{P} = \{ p \} $ containing all $N \times N$ patches in $[1..W] \times [1..H]$ can be obtained. Then, the repeatability loss is defined as:
\begin{equation}
\begin{split}
\mathcal{L}_{rep}(I,I',U) = \mathcal{L}_{cos}(I,I',U) + \mathcal{L}_{peaky}(I) \\ + \mathcal{L}_{peaky}(I')
\end{split}
\end{equation}
where $\mathcal{L}_{cos}$ and $ \mathcal{L}_{peaky}$ is defined as
\begin{equation}
\mathcal{L}_{cos}(I,I',U) = 1- \frac{1}{ |\mathcal{P}| } \sum_{p \in \mathcal{P}}( S[p] ,S'_U[p])
\end{equation}
\begin{equation}
\mathcal{L}_{peaky}(I) = 1- \frac{1}{ |\mathcal{P}| } \sum_{p \in \mathcal{P}}( \max \limits_{i,j \in p} S_{ij} - 
\mathop{mean} \limits_{i,j \in p} S_{ij} )
\end{equation}

Next, the Average Precision (AP) loss is used to learn the reliability in R2D2:
\begin{equation}
\mathcal{L}_{AP \kappa}(i,j) = 1- [AP(i,j) R_{ij} + \kappa (1 - R_{ij})]
\end{equation}
where $ \kappa \in [0,1]$ is a hyperparameter, which indicates the minimum expected AP for each patch. $\lambda$ is also a hyperparameter to balance the initial prediction loss $\mathcal{L}_{initial}$ and the final prediction loss $\mathcal{L}_{final}$.
\begin{equation}
\mathcal{L}_{total} = \mathcal{L}_{initial} + \lambda \mathcal{L}_{final}
\end{equation}

\subsection{Results} 
\begin{figure}[t]
	\begin{center}
		\includegraphics[width=0.5\textwidth]{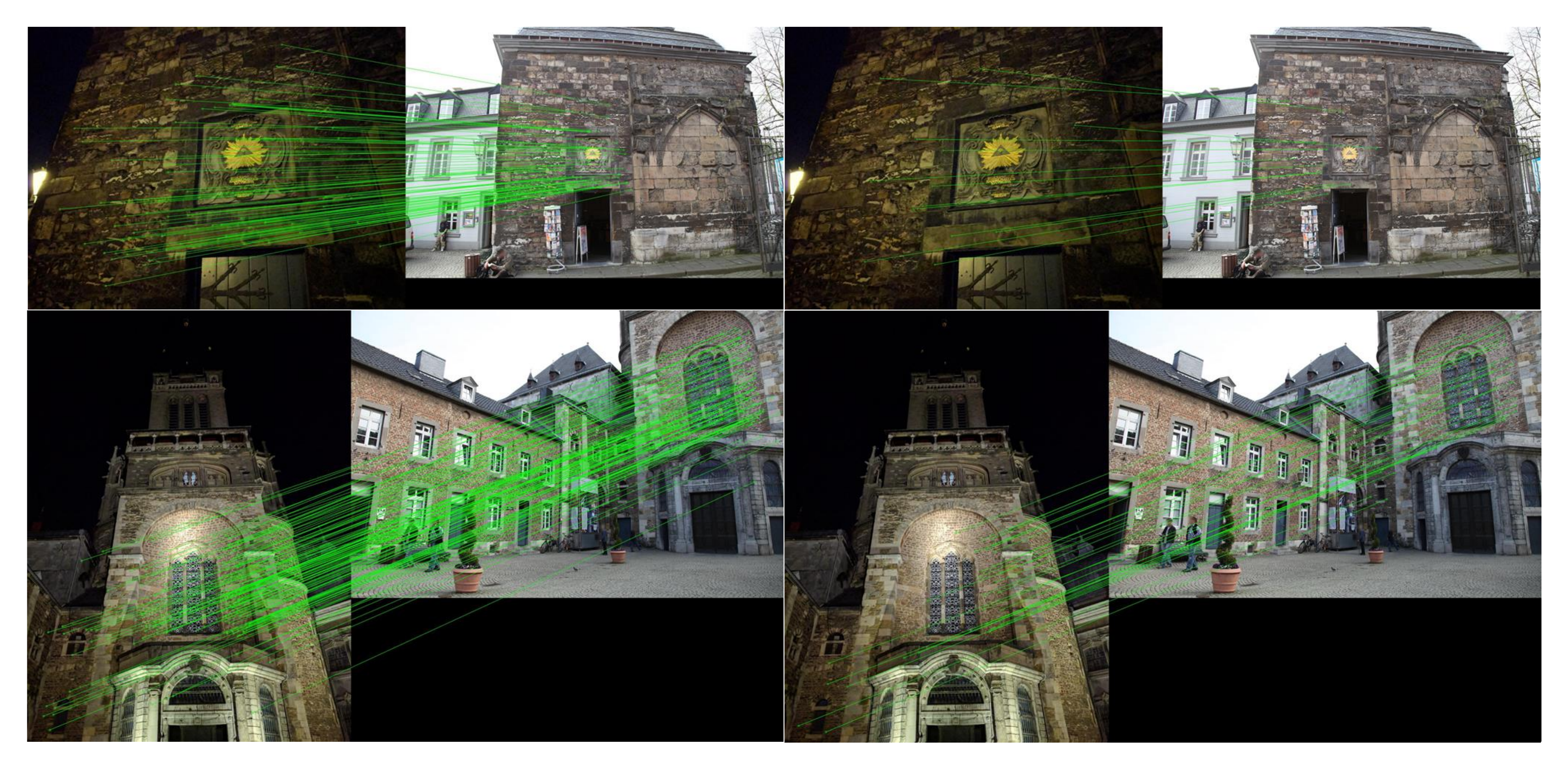}
	\end{center}
	\caption{Examples of matches obtained by our method (1st col) and the original R2D2 method (2nd col).}
	\label{fig:cmp}
\end{figure}

We compare our method to previous methods in recent years in Table~\ref{table1}. Our method achieves the state-of-the-art performance in two metrics. As shown in Figure~\ref{fig:cmp}, our method detects more robust keypoints than the original R2D2 method based on KNN matcher.

We conducted several experiments on the dataset to justify our design modules, the results are shown in Table~\ref{table2}. With the local-guided module, the performance on $(1m, 5^{\circ})$ is improved from 65.3 to 67.3. Further, the full model with both local-guided and multi-scale aggregation modules obtains better performance in terms of $(0.5m, 2^{\circ})$ and $(5m, 10^{\circ})$. These results demonstrate that introducing high-level semantic information into low-level textual features is beneficial to learn robust local features.

\begin{table}
\begin{center}
\begin{tabular}{|l|c|c|c|}
			\hline
			Method & $(0.5m, 2^{\circ})$ & $(1m, 5^{\circ})$ & $(5m, 10^{\circ})$\\
			\hline
			RootSIFT~\cite{lowe2004distinctive} & 33.7 & 52.0 &  65.3 \\
			HAN + HN++~\cite{mishkin2018repeatability} & 37.8 & 54.1 &  75.5\\
			SuperPoint~\cite{detone2018superpoint} & 42.8 & 57.1 &  75.5 \\
			R2D2~\cite{revaud2019r2d2} & 45.9 & 66.3 &  \textbf{88.8} \\
			D2-Net~\cite{dusmanu2019d2} & 45.9 & 67.3 & \textbf{88.8} \\
			ASLFeat(10K)~\cite{luo2020aslfeat} & 46.9 & 65.3 & \textbf{88.8} \\
			UR2KID~\cite{yang2020ur2kid} & 46.9 & 67.3 & \textbf{88.8} \\
			\hline
			R2D2(40K) & \textbf{48.0} & 67.3 & \textbf{88.8} \\
			ASLFeat(10K) + OANet &  \textbf{48.0} & 67.3 & \textbf{88.8} \\
			SuperPoint + SuperGlue & 45.9 &	70.4 & \textbf{88.8} \\
			ONavi & \textbf{48.0} & \textbf{71.4} &  \textbf{88.8} \\
			\hline
			Ours(20K) &  \textbf{48.0} & 65.3 & \textbf{88.8} \\
			\hline
\end{tabular}
\end{center}
\caption{Evaluation results on the Aachen Day-Night dataset for visual localization.}
\label{table1}
\end{table}

\begin{table}
\begin{center}
\begin{tabular}{|l|c|c|c|}
			\hline
			Method & $(0.5m, 2^{\circ})$ & $(1m, 5^{\circ})$ & $(5m, 10^{\circ})$\\
			\hline
			R2D2 (baseline) & 45.9 & 65.3 &  87.8 \\
            R2D2 + LG  & 45.9 &  \textbf{67.3} &  87.8 \\
            R2D2 + LG + MA  &  \textbf{48.0} & 65.3 &   \textbf{88.8} \\
			\hline
		\end{tabular}
\end{center}
\caption{Ablation study of our network on the Aachen Day-Night dataset, where + LG and + MA represents the method with the local-guided module and the multi-scale module.}
\label{table2}
\end{table}

\section{Conclusion}
In this paper, we adopt R2D2 and HRNet as our backbone to jointly learn local features detectors and descriptors. The local-guided module and the multi-scale aggregation module can successfully fuse low-level textual features and high-level semantic features to improve the robustness of local features. Experimental results on a challenging dataset show that our method achieves the state-of-the-art performance in two metrics.

{\small
\bibliographystyle{ieee_fullname}
\bibliography{egbib}
}

\end{document}